\documentclass[journal]{IEEEtran}
\ifCLASSINFOpdf
\else
\fi

\usepackage{booktabs}
\usepackage{amsfonts}
\usepackage{graphicx}
\usepackage{multirow}
\usepackage{algorithm}
\usepackage{algpseudocode}
\usepackage{amsmath}
\usepackage{caption2}

\usepackage{color}

\newtheorem{myDef}{Definition}

\begin{document}
%
\title{Language-Level Semantics Conditioned 3D Point Cloud Segmentation}
%
%
%

\author{Bo Liu, Shuang Deng, Qiulei Dong, and Zhanyi Hu
\thanks{B. Liu and Z. Hu are with the National Laboratory of Pattern Recognition, Institute of Automation, Chinese Academy of Sciences, Beijing 100190, China, and also with the School of Future Technology, University of Chinese Academy of Sciences, Beijing 100049, China (e-mail: liubo2017@ia.ac.cn; huzy@nlpr.ia.ac.cn).}
\thanks{S. Deng and Q. Dong are with the National Laboratory of Pattern Recognition, Institute of Automation, Chinese Academy of Sciences, Beijing 100190, China, and also with the School of Artificial Intelligence, University of Chinese Academy of Sciences, Beijing 100049, China. Q. Dong is also with the Center for Excellence in Brain Science and Intelligence Technology, Chinese Academy of Sciences, Beijing 100190, China (e-mail: shuang.deng@nlpr.ia.ac.cn; qldong@nlpr.ia.ac.cn).
\textit{(Corresponding author: Qiulei Dong.)}}}

%
%

\markboth{}%
{Shell \MakeLowercase{\textit{et al.}}: Bare Demo of IEEEtran.cls for IEEE Journals}
%



\maketitle

\begin{abstract}
In this work, a language-level \textit{Se}mantics \textit{Cond}itioned framework for 3D \textit{Point} cloud segmentation, called \textit{SeCondPoint}, is proposed, where language-level semantics are introduced to condition the modeling of point feature distribution as well as the pseudo-feature generation, and a feature-geometry-based mixup approach is further proposed to facilitate the distribution learning. To our knowledge, this is the first attempt in literature to introduce language-level semantics to the 3D point cloud segmentation task. Since a large number of point features could be generated from the learned distribution thanks to the semantics conditioned modeling, any existing segmentation network could be embedded into the proposed framework to boost its performance. In addition, the proposed framework has the inherent advantage of dealing with novel classes, which seems an impossible feat for the current segmentation networks. Extensive experimental results on two public datasets demonstrate that three typical segmentation networks could achieve significant improvements over their original performances after enhancement by the proposed framework in the conventional 3D segmentation task. Two benchmarks are also introduced for a newly introduced zero-shot 3D segmentation task, and the results also validate the proposed framework.
\end{abstract}


%
\IEEEpeerreviewmaketitle

\section{Introduction}
\IEEEPARstart{P}{oint} cloud semantic segmentation is a fundamental problem in the computer vision and computer graphics communities. Inspired by the tremendous success of deep neural networks (DNNs) in the 2D image analysis field, recently, DNNs have been introduced to various 3D point cloud processing tasks~\cite{2015voxnet,2017pointnet, deng2021rotation}. However, due to the irregular structure of 3D point clouds, the existing DNNs which were usually designed to work at regular grid inputs (e.g. image pixels), could not straightforwardly be applied to 3D point clouds.

To tackle this problem, a large number of DNN-based methods~\cite{Hu2020JSENetJS,Mao2019InterpolatedCN,Jiang2019HierarchicalPI} have been proposed to design new network architectures for 3D point cloud segmentation, which could be roughly divided into two categories: projection-based methods and point-based methods. Projection-based methods~\cite{2015voxnet,2017dpss,2018squeezeseg} projected irregular 3D point clouds into regular 3D occupancy voxels or 2D image pixels, where conventional 3D or 2D convolutional neural networks (CNNs) could be applied directly. Despite their decent performances, the projection-based methods usually suffer from information loss in the projection process. Recently, many point-based methods~\cite{2017pointnet,2020randlanet,2019kpconv, 2020dpc} have been proposed. The point-based methods directly processed 3D points by extracting point features with multi-layer perceptrons (MLPs) and then aggregating neighborhood point features via a specially designed point feature aggregator or an elaborate convolution operation. Since the point-based methods directly operate on 3D points, the key to the point-based methods is how to successfully aggregate local and long-range geometrical information. Currently, due to the lack of large-scale datasets (like ImageNet for 2D image analysis) in the 3D point cloud processing field, both projection-based and point-based methods generally suffer from the data hungry problem due to the adoption of DNN architectures.

Data hungry, on the one hand, could be alleviated by data augmentation techniques. Some works have been proposed to augment 3D point clouds. Hand-crafted rules (e.g. random rotating and jittering) were used as general pre-processing operations by many works~\cite{2017pointnet,2017pointnet++}. PointAugment~\cite{Li2020PointAugment} introduced an auto-augmentation framework, which took an adversarial learning strategy to jointly optimize a point cloud augmentation network and a classification network. Following the Mixup~\cite{ZhangCDL18}, an effective augmentation technique in the 2D image analysis field, PointMixup~\cite{chen2020pointmixup} and RSMix~\cite{lee2021regularization} mixed samples in the 3D space to augment 3D point clouds. However, these approaches focused on point-cloud-level augmentation and only aimed at the point cloud classification task, and they are generally not quite effective for the point cloud segmentation task because point cloud segmentation is a point-level discrimination task.

Data hungry in the 3D segmentation task, on the other hand, is reflected by the fact that novel object classes usually appear in real 3D scenes, and it is impossible for human annotators to annotate all potential object classes for model learning. To tackle this novel-class problem, we introduce a new task by generalizing zero-shot learning to 3D semantic segmentation in this work, called zero-shot 3D scene semantic segmentation where some novel 3D object classes (no training data available) need to be classified in the testing 3D scenes. Note that although Cheraghian et al.~\cite{cheraghian2020transductive} introduced zero-shot learning to 3D shape classification to recognize novel-class 3D object shapes, zero-shot 3D scene segmentation is considerably different from zero-shot 3D shape classification, because 1) feature learning of seen-class objects and novel-class objects are affected mutually in the local feature aggregation process in the 3D scene segmentation task, while feature learning of different classes in the 3D shape classification task is totally independent since all points in a single 3D object shape belong to the same class; 2) 3D scenes are naturally and structurally class-level imbalanced.

\begin{figure}
	\centering
	\includegraphics[width=0.98\columnwidth]{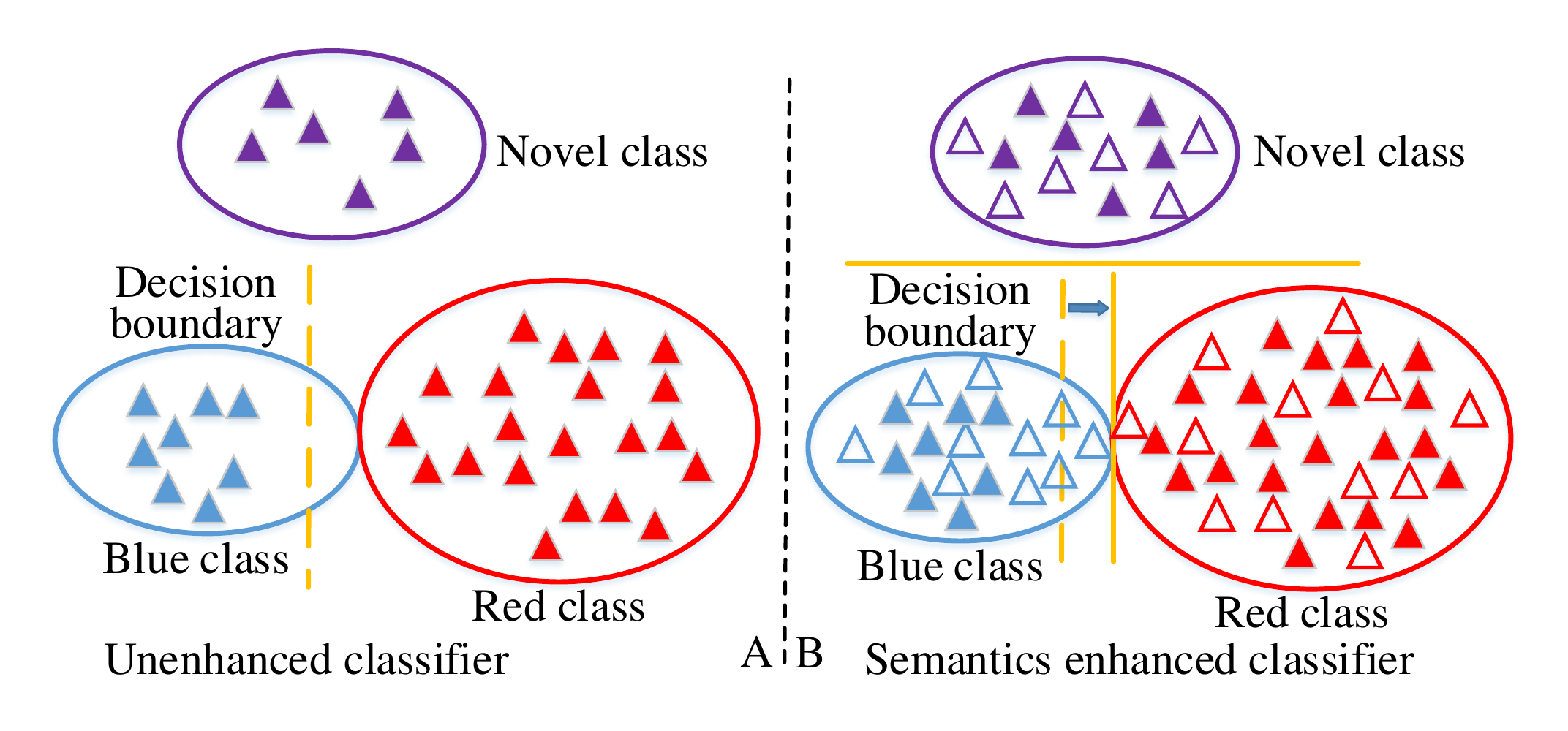}
	\caption{Blue/red/purple ellipses represent the real data distributions of blue/red/novel classes respectively.  Solid blue/red triangles represent samples from seen blue/red classes in the training set and solid purple triangles represent novel-class samples which are not included in the training set. Hollow blue/red/purple triangles represent samples of blue/red/novel classes generated by our SeCondPoint. Yellow dashed line represents the decision boundary determined by the unenhanced classifier, and yellow solid line represents the decision boundary determined by the enhanced classifier. In A, the decision boundary of the unenhanced classifier locates within the distribution of blue class, and the novel class could not be discriminated from the other two classes. In B, the semantics enhanced classifier is enhanced in two aspects: 1) the decision boundary between blue and red classes is pushed close to the real boundary, bringing better generalization to classify testing samples; 2) the semantics enhanced classifier can classify novel class.}
	\label{fig1}
\end{figure}

In this work, we propose a unified framework called SeCondPoint for both conventional and zero-shot 3D point cloud segmentation, where language-level semantics are introduced to condition the modeling of point feature distribution as well as the pseudo-feature generation. The proposed SeCondPoint firstly employs a conditional generative adversarial network to model point feature distribution conditioned on semantic information, which is adversarially learned from the real distribution of point features extracted by an arbitrary existing point cloud segmentation network. A feature-geometry-based mixup approach is further proposed to facilitate the distribution learning. After training the generative model, a large number of point features could be generated from the learned distribution conditioned on the specific semantic information. With these generated point features, a semantics enhanced point feature classifier can be trained for point cloud segmentation. As illustrated in Figure~\ref{fig1}, the advantages of the semantics enhanced classifier include: 1) for a class with a relatively small number of point samples in the training 3D point clouds, a great number of point features could be generated from the learned distribution conditioned on the corresponding semantic information of this class. Hence, the decision boundary between this class and its neighboring class in the feature space could be improved to be better generalized to the testing samples; 2) due to the semantics conditioned modeling, point features of novel classes could be generated conditioned on their corresponding semantic information, which endows the learned classifier with the ability to segment novel-class objects.

In sum, the key contributions of this work include:
\begin{itemize}
	\item We propose a unified framework (SeCondPoint) for both conventional 3D scene semantic segmentation (C3DS) and zero-shot 3D scene semantic segmentation (Z3DS), where language-level semantics are introduced to conditionally model point feature distribution and generate point features. Any existing C3DS networks could be seamlessly embedded into the proposed framework for either improving their C3DS performances or classifying novel-class objects. To our best knowledge, this is the first attempt to utilize language-level semantics for 3D scene semantic segmentation in this field.
	\item Under the framework of zero-shot learning, we introduce the new task of Z3DS (zero-shot 3D scene semantic segmentation) and construct two benchmarks for algorithmic evaluation. Two baselines are also established for comparison. To our best knowledge, this is the first such an attempt to investigate the Z3DS problem, and the introduced two benchmarks and methodology could be of help for more researches on this new direction.
	\item Extensive experimental results in Section~\ref{experiments} show that the proposed framework is not only able to significantly improve the C3DS performances of existing segmentation networks, but also establish a competitive baseline in the Z3DS task.
\end{itemize}

The remaining of this paper is organized as follows. Firstly, we review some related works in Section~\ref{related}. Secondly, we elaborate the proposed framework in Section~\ref{methodology}, where a language-level semantics conditioned feature modeling network and a semantics enhanced feature classifier are discussed. In Section~\ref{experiments}, the experimental setup, extensive results, and some in-depth discussions are provided. Finally, we conclude the paper and outline some future works in Section~\ref{conclusion}.

\section{Related work}
\label{related}

\subsection{Deep Learning for Point Cloud Segmentation}
As discussed in Section~\uppercase\expandafter{\romannumeral1}, the existing DNN-based methods for point cloud segmentation could be roughly classified into projection-based methods and point-based methods.\\
\textbf{Projection-based methods}: Multi-view based method~\cite{2017dpss} projected 3D point clouds onto 2D planes from multiple virtual camera views. Similarly, spherical representation based method~\cite{2018squeezeseg} projected a 3D point cloud onto a 2D sphere image. 3D convolution-based methods~\cite{2015voxnet, 2018sscn, 2019minko,Zhang2020PolarNetAI} divided 3D point clouds into multiple regular sets of occupancy voxels and then elaborated effective convolution operations on them. However, the performances of such projection-based methods are severely affected by the resolution of regular data and 3D object occlusion.\\
\textbf{Point-based methods}: PointNet~\cite{2017pointnet} is the pioneering work to directly process 3D point clouds using shared MLPs and max-pooling layers. Based on PointNet, a vast number of point-based networks have been proposed. These methods could be further divided into point-wise MLP methods, point convolution methods, and graph-based methods. Point-wise MLP methods~\cite{2017pointnet++, 2018pointsift, 2019sagss, 2020randlanet, 2020point2node,2020pointgcr,2020pointnl,Liu2020ACL,deng2021ga,deng2021superpoint} used MLPs as the basic unit and implemented different local aggregation operations for capturing local geometry of 3D point clouds. Point convolution methods~\cite{2018pointcnn, 2018pcnn, 2018spidercnn, 2019kpconv, 2020dpc,2019acnn,2020fkaconv,Lei2020SegGCNE3,Liu2019RelationShapeCN,Lei2019OctreeGC} implemented effective convolution operations to transform and aggregate features for 3D point clouds. Graph-based methods~\cite{2018spgraph, 2019pointweb, 2018dgcnn, 2019gacnet, He2019GeoNetDG} treated 3D point clouds as a graph structure and used graph convolution operations to capture underlying shapes and geometric structures of 3D point clouds.

\subsection{Data Augmentation for Point Clouds}
Hand-crafted rule based augmentations (e.g. random rotating and jittering) are usually used to augment 3D point clouds by many existing methods~\cite{2017pointnet,2017pointnet++,2018dgcnn}. Recently, Li et al.~\cite{Li2020PointAugment} proposed the PointAugment framework to automatically augment point clouds, where a point cloud augmentation network was used to augment point clouds and a classification network was adopted to classify the augmented point clouds. By training the two networks adversarially, the learned classification network was expected to be enhanced. Mixup~\cite{ZhangCDL18} has achieved a huge success in the 2D image analysis field, but it could not be used directly in 3D point clouds due to the irregular structure of point clouds. Addressing this problem, PointMixup~\cite{chen2020pointmixup} mixed different parts of two point clouds by solving an optimal transport problem to augment point clouds. To reduce the loss of structural information of the original samples in PointMixup, Lee et al.~\cite{lee2021regularization} proposed a shape-preserving augmentation technique, called RSMix which could partially mix two point clouds while preserving the partial shape of the original point clouds. Our proposed framework also aims to augment training data, however it is significantly different from these related works mainly in three aspects: 1) the related works are designed only for the 3D shape classification task by point-cloud-level augmentation, while ours is proposed for the 3D scene segmentation task by point-level feature distribution modeling. Considering that 3D scenes usually include millions of 3D points and such methods proceed based on complex optimization on 3D points, it is hard for them to adapt to 3D segmentation scenario; 2) 3D scene segmentation faces essentially different problems from 3D shape classification in terms of feature learning and inherent class imbalance; 3) our proposed method significantly differs from the related works in its working principle. 

\subsection{Zero-Shot 2D Image Semantic Segmentation}
Zero-shot image semantic segmentation~\cite{bucher2019zero,GuZ00Z20,XianC0SA19} has received increasing attention only recently. Xian et al.~\cite{XianC0SA19} proposed a semantic projection network to perform semantic segmentation of novel classes, where visual features were projected into a semantic feature space and then novel classes were classified according to similarities between the projected features and novel-class semantic features. Both Bucher et al.~\cite{bucher2019zero} and Gu et al.~\cite{GuZ00Z20} proposed similar methods which combined a deep visual segmentation model with a generative model to generate visual features from semantic features and trained a visual feature classifier to segment images. Compared with such zero-shot 2D image learning tasks, our introduced Z3DS task is in fact a more challenging task because 1) for regular 2D images, CNN is powerful to extract discriminative features. However, there seems no such powerful architectures for irregular 3D point clouds; 2) point feature extraction is mainly based on geometrical information, but some well-defined semantic features which could well align with geometrical features for zero-shot 3D point inference, are usually lacking.

\subsection{Generative Models for Point Clouds}
Recently, generative models like GAN~\cite{Goodfellow2014GenerativeAN} (Generative Adversarial Network), VAE~\cite{KingmaW13} (Variational Auto-Encoder), and NF~\cite{Rezende2015} (Normalizing Flows) have been applied to point cloud generation. Achlioptas et al.~\cite{AchlioptasDMG18} introduced GAN to point cloud generation, while Gadelha et al.~\cite{GadelhaWM18} and Zamorski et al.~\cite{Zamorski2018} employed VAE to generate point clouds. In such works, point cloud was modeled with a point distribution and was synthesized by sampling points from the learned distribution. Yang et al.~\cite{YangHH0BH19} applied NF to point cloud generation by learning a two-level hierarchy of distributions where the first level was the distribution of shapes and the second level was the distribution of points. Unlike these methods, our method learns point feature distribution conditioned on the semantics for each semantic class, which is then used to train a discriminant model by sampling a large number of point features from the learned distribution for the point cloud segmentation task.

\begin{figure*}
	\centering
	\includegraphics[width=1.6\columnwidth]{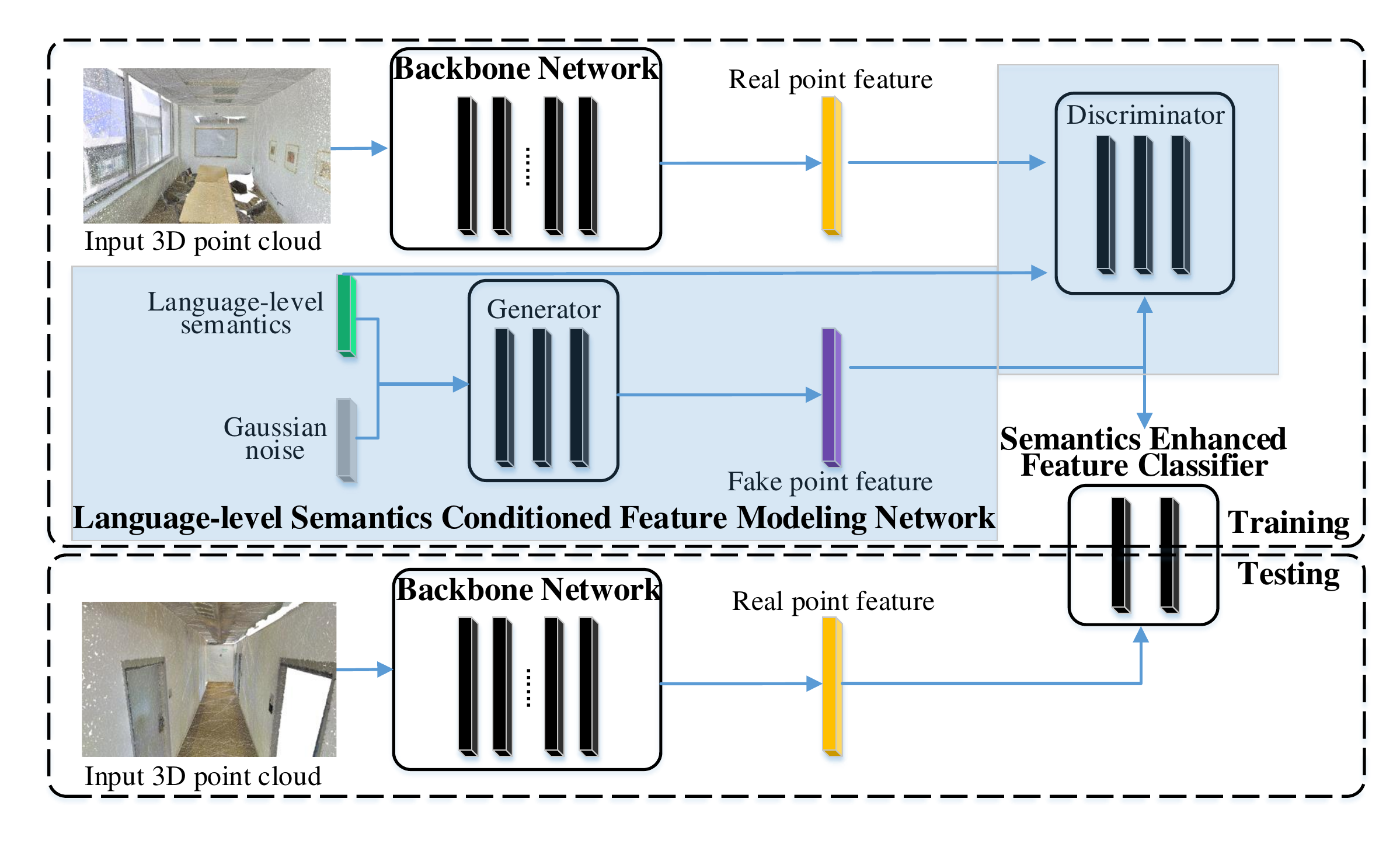}
	\caption{The proposed SeCondPoint framework consists of three parts: a pre-trained backbone segmentation network for extracting point features from input 3D point clouds; a language-level semantics conditioned feature modeling network for learning point feature distributions conditioned on the corresponding semantic features; a semantics enhanced feature classifier for classifying both seen-class and unseen-class points in 3D point clouds. Note that the parameters of the pre-trained backbone network are fixed while those of the feature modeling network and feature classifier need to be learned.}
	\label{fig2}
\end{figure*}

\section{Methodology}
\label{methodology}
We begin by introducing the task of conventional 3D scene semantic segmentation (C3DS), zero-shot 3D scene semantic segmentation (Z3DS), and some necessary notations. Suppose we are given a 3D point cloud scene set $\mathcal{P}$ at the training stage, for the sake of clarity, which can also be represented as a 3D point set $\mathcal{D}^{S}_{tr}=\{(\mathbf{x}_{n},y_{n})\}_{n=1}^{N}$, where $\mathbf{x}_{n}$ is a point in a 3D point cloud scene and $y_{n}$ is the label of $\mathbf{x}_{n}$, belonging to the seen-class label set $Y^{S}$, and $N$ is the number of points in all point clouds. Usually, $\mathbf{x}_{n}$ is a $(3+a)$-D feature vector with $3$-D X-Y-Z Euclidean coordinates and $a$-D features (e.g. $3$-D R-G-B color features). In the deep learning community, a feature extractor $u(\cdot)$ is usually used to transform the original $(3+a)$-D point feature vectors into more high-level point feature vectors. Hence in this paper, $\mathbf{x}_{n}$ is used to refer to the feature vector after the processing of the feature extractor $u(\cdot)$. Suppose we are additionally given a semantic feature set $\mathcal{E}=\{ \mathbf{e}_{y} \mid y \in Y \}$, where $\mathbf{e}_{y}$ is the semantic feature of class $y$, and $Y$ is a class label set which not only includes the seen-class label set $Y^{S}$ but also includes the unseen-class label set $Y^{U}$.  

At the testing stage, the task is to segment the testing 3D point cloud scenes by classifying each point in the 3D point set $\mathcal{D}_{te}=\{\mathbf{x}_{m}\}_{m=1}^{M}$, where $M$ is the number of points in $\mathcal{D}_{te}$.
\begin{myDef}
	C3DS: Suppose the labels of the points in $\mathcal{D}_{te}$ belong to $Y^{S}$, given $\mathcal{D}^{S}_{tr}$ and $\mathcal{E}$, the goal is to learn a mapping $f:\mathcal{D}_{te} \to Y^{S}$.
\end{myDef}
\begin{myDef}
	Z3DS: Suppose the labels of the points in $\mathcal{D}_{te}$ belong to $Y^{U}$, given $\mathcal{D}^{S}_{tr}$ and $\mathcal{E}$, the goal is to learn a mapping $f:\mathcal{D}_{te} \to Y^{U}$.
\end{myDef}
\begin{myDef}
	Generalized Z3DS (GZ3DS): Suppose the labels of the points in $\mathcal{D}_{te}$ belong to $Y$, given $\mathcal{D}^{S}_{tr}$ and $\mathcal{E}$, the goal is to learn a mapping $f:\mathcal{D}_{te} \to Y$.
\end{myDef}
Note that the 3D scene semantic segmentation problem is actually a point-level classification problem, hence we could represent point cloud set as corresponding point set and segment 3D scenes by classifying 3D points. It is assumed in C3DS that testing points only belong to the seen-class label set, this is to say, C3DS is unable to tackle novel-class objects often occurred in realistic scenarios. Z3DS assumes that we have the prior that testing points belong to only the unseen-class label set. Although this prior is not always available in practice, Z3DS can demonstrate the ability of a model to infer unseen-class objects, and is of a practical value with the aid of some additional modules (like novel-class detector). GZ3DS simultaneously deals with seen classes and unseen classes, which is a more practical setting.

Here, we propose the SeCondPoint framework for 3D scene semantic segmentation, which utilizes language-level semantics to condition the modeling of point feature distribution and the point feature generation. As shown in Figure \ref{fig2}, the proposed SeCondPoint consists of three parts: a backbone segmentation network for extracting point features from input 3D point clouds, a language-level semantics conditioned feature modeling network for learning point feature distribution, and a semantics enhanced feature classifier for classifying both seen-class and unseen-class points in 3D point clouds for semantic segmentation. Here we would point out that an arbitrary existing (also novel) segmentation network could be used as backbone network under the SeCondPoint framework. Since our main goal is to propose the novel SeCondPoint framework, rather than a novel segmentation network, we simply use existing segmentation networks as backbone networks here. In the following, we describe the language-level semantics conditioned feature modeling network and the semantics enhanced feature classifier respectively. Note that in the proposed SeCondPoint framework, the parameters of the backbone network are fixed, but the remaining two modules need to be learned.

\subsection{Language-Level Semantics Conditioned Feature Modeling Network}
\subsubsection{Adversarial Feature Modeling}
Here, we propose a language-level semantics conditioned feature modeling network to learn the conditional distribution of point features conditioned on semantic features by introducing language-level semantic information of both seen and unseen (novel) object classes. Note that there exist many models for extracting language-level semantics in literature~\cite{PenningtonSM14,MikolovSCCD13,Devlin2019BERTPO}, here we use the semantic embeddings of the object class names extracted by the existing language model~\cite{MikolovSCCD13}, considering that our current work is to show the usefulness of semantics-conditioned feature modeling in 3D scene segmentation, rather than comparing different semantics. 

As shown in Figure~\ref{fig2}, the feature modeling network employs a conditional generative adversarial network to model the conditional distributions, where the inputs of the generator are the concatenation of semantic features of object classes and Gaussian noises, the outputs of the generator are synthesized point features of the corresponding object classes, and the discriminator is to discriminate the real point features extracted by the backbone network from those generated by the generator. Specifically, suppose we are given a backbone network $u(\cdot)$, a generator $G(\cdot)$, and a discriminator $D(\cdot)$. For a given input 3D point with its corresponding class label $y$, we firstly extract point feature $\mathbf{x} \in \mathbb{R}^{b}$ of the input 3D point with the backbone network $u(\cdot)$. Then, the generator $G(\cdot)$ is used to generate a fake point feature $\bar{\mathbf{x}}$ conditioned on the corresponding semantic feature $\mathbf{e}_{y}$, and the discriminator $D(\cdot)$ is to discriminate the generated fake point feature $\bar{\mathbf{x}}$ from the real point feature $\mathbf{x}$. In order to learn a point feature distribution which could not be discriminated from the real one by the discriminator, the generator is adversarially trained with the discriminator under the framework of Wasserstein Generative Adversarial Network (WGAN)~\cite{arjovsky2017wgan} as follows:
\begin{equation}
	\begin{split}
		L_{wgan} = & \min_{G} \max_{D}  E_{(\mathbf{x},y) \in \mathcal{D}_{tr}^{S}} [D(\mathbf{x},\mathbf{e}_{y})] - E_{y \in \mathcal{D}_{tr}^{S}} [D(\bar{\mathbf{x}},\mathbf{e}_{y})] \\
		& - \lambda E_{(\mathbf{x},y) \in \mathcal{D}_{tr}^{S}} [(||\nabla_{\hat{\mathbf{x}}} D(\hat{\mathbf{x}},\mathbf{e}_{y})||_{2}-1)^{2}]
	\end{split}
	\label{eq1}
\end{equation}
where $\mathcal{D}_{tr}^{S}$ is the labeled training point feature set extracted by the backbone network from the training 3D point clouds. The first two terms in (\ref{eq1}) are the original objectives of WGAN, which aim to minimize the Wasserstein distance between the distributions of the generated point features and those of the real point features, $\bar{\mathbf{x}}$ is generated by the generator $G(\cdot)$ conditioned on the corresponding semantic feature $\mathbf{e}_{y}$ and a standard Gaussian noise $\mathbf{z} \sim \mathcal{N}(\mathbf{0},\mathbf{I})$, i.e. $\bar{\mathbf{x}} = G(\mathbf{e}_{y},\mathbf{z})$. The third term in (\ref{eq1}) is used for gradient penalty for the discriminator, where $\hat{\mathbf{x}} = \alpha \mathbf{x} + (1-\alpha) \bar{\mathbf{x}}$ with $\alpha$ sampled from a uniform distribution, i.e. $\alpha \sim \mathcal{U}(0,1)$, is used to estimate gradients, and $\lambda$ is a hyper-parameter for weighting the gradient penalty term, which is usually set to $10$ as suggested in~\cite{arjovsky2017wgan}. By optimizing the MinMax objective, the generator can finally learn the conditional distribution of real point features of each class conditioned on its semantic feature. In other words, an arbitrary number of point features for each class could be synthesized by sampling features from the learned conditional distribution.

\subsubsection{Feature-Geometry-Based Mixup Training}
Feature learning in 3D scene semantic segmentation is mainly based on geometrical information. This is to say, geometrically adjacent points generally have similar features in the learned feature space. At the same time, due to the feature similarity between geometrically adjacent object classes, confusing classifications often happen between these classes. Inspired by Mixup~\cite{ZhangCDL18}, here we propose a feature-geometry-based mixup training approach to increase the discrimination between adjacent object classes in the feature space. Specifically, we firstly compute the class feature centers $\{\mathbf{X}_{0}, \cdots, \mathbf{X}_{c}, \cdots, \mathbf{X}_{C}\}$, where $\mathbf{X}_{c} = \frac{1}{n_{c}} \sum_{i=1}^{n_{c}} \mathbf{x}_{i}^{c}$, $n_{c}$ is the number of points belonging to class $c$, and $C$ is the number of classes. Then we compute the Euclidean similarity matrix $A$ between these feature centers. Next, given a point feature $\mathbf{x}^{c}$ from the class $c$, we firstly identify the closest $I$ classes with the class $c$ according to the similarity matrix $A$, denoted by $\bar{c}$, and then we sample a point feature from $\bar{c}$, denoted by $\mathbf{x}^{\bar{c}}$. Finally, an intermediate feature sample $\mathbf{x}$ is synthesized by interpolating between $\mathbf{x}^{c}$ and $\mathbf{x}^{\bar{c}}$ with a scalar $\beta$ sampled from the Uniform distribution $\mathcal{U}(0,1)$ as:
\begin{equation}
	\begin{split}
		\mathbf{x} = & \beta * \mathbf{x}^{c} + (1-\beta) * \mathbf{x}^{\bar{c}} \\
		\mathbf{e} = & \beta * \mathbf{e}^{c} + (1-\beta) * \mathbf{e}^{\bar{c}} \\
	\end{split}
	\label{eq2}
\end{equation}
where $\mathbf{e}^{c}$ and $\mathbf{e}^{\bar{c}}$ are the corresponding semantic features of $\mathbf{x}^{c}$ and $\mathbf{x}^{\bar{c}}$. According to (\ref{eq2}), we can interpolate a large number of point features which locate between two geometrically adjacent classes in the feature space. Finally, we add these interpolated samples to the training set to train the feature generative modeling network. Here we use a hyper-parameter $\gamma$ to control the scale of interpolated samples, which is defined as the ratio of the interpolated samples to the real samples.

\subsection{Semantics Enhanced Feature Classifier}
Once the language-level semantics conditioned feature modeling network is trained, a large number ($K$) of point features for each object class could be generated conditioned on the corresponding semantic feature $\mathbf{e}_{y}$ and $K$ different random noises $\{\mathbf{z}_{k}\}_{k=1}^{K}$ sampled from the standard Gaussian distribution $\mathcal{N}(\mathbf{0},\mathbf{I})$. Specifically, we generate point features according to:
\begin{equation}
	\bar{\mathbf{x}} = G(\mathbf{e}_{y}, \mathbf{z})
	\label{eq3}
\end{equation}
where $\bar{\mathbf{x}}$ is the generated point feature. In the following, we describe the feature generation and classifier learning in three different tasks, i.e. C3DS, Z3DS, and GZ3DS.

\subsubsection{Conventional 3D Scene Semantic Segmentation}
In C3DS, the testing points are assumed to come from only seen classes, hence, we generate a large number of point features for each seen class in $Y^{S}$ conditioned on seen-class semantic features according to (\ref{eq3}). The set of generated point features and corresponding labels is denoted by $\bar{\mathcal{X}}^{S}$. Then, we train a semantics enhanced classifier $f_{s}(\cdot)$ with $\bar{\mathcal{X}}^{S}$ as follows:
\begin{equation}
	L_{s} = E_{(\bar{\mathbf{x}},y) \in \bar{\mathcal{X}}^{S}} [\mathbb{C}(\mathbb{S}(f_{s}(\bar{\mathbf{x}})),y)]
	\label{eq4}
\end{equation}
where $\mathbb{C}(\cdot)$ and $\mathbb{S}(\cdot)$ are a cross-entropy loss function and a softmax function respectively. Note that $f_{s}(\cdot)$ could be any classifier, not constrained to linear classifier as done in existing segmentation networks~\cite{2018dgcnn, 2020randlanet, fan2021scf}. After training the classifier, given a real testing point feature $\mathbf{x}$, we predict its label $y$ by:
\begin{equation}
	y = \arg \max_{y \in \mathcal{Y^{S}}} \mathbb{S}(f_{s}(\mathbf{x}))
	\label{eq5}
\end{equation}
Finally, for a given testing 3D point cloud, we achieve point cloud semantic segmentation by classifying the feature of each 3D point extracted by the backbone network via the learned semantics enhanced feature classifier.

\subsubsection{Zero-Shot 3D Scene Semantic Segmentation}
Thanks to the language-level semantics conditioned point feature modeling, the proposed framework has the flexible ability to segment novel-class objects in 3D point cloud scenes (including Z3DS and GZ3DS) if their corresponding semantics are available, which is an impossible function for existing segmentation networks. In Z3DS, the task is to classify unseen-class points. To this end, we sample a large number of point features for each unseen class in $Y^{U}$ from the learned conditional distributions conditioned on the unseen-class semantic features according to (\ref{eq3}). That's to say, the semantic feature $\mathbf{e}_{y}$ is sampled from unseen-class label set $Y^{U}$, which is different from those in C3DS where the semantic feature $\mathbf{e}_{y}$ is sampled from seen-class label set $Y^{S}$. Then we train a semantics enhanced classifier $f_{u}(\cdot)$ in a similar way to (\ref{eq4}), and classify the real testing unseen-class points as done in (\ref{eq5}). 

\subsubsection{Generalized Zero-Shot 3D Scene Semantic Segmentation}
In GZ3DS, the testing points could be from either seen classes or unseen classes. Hence, according to (\ref{eq3}), we generate a large number of point features for every class in $Y$ conditioned on all semantic features, and the semantics enhanced classifier $f_{g}(\cdot)$ training and the point feature classifying are similar to those in C3DS and Z3DS. The only difference among C3DS, Z3DS and GZ3DS is that their conditions and classification space are different, which in turn demonstrates the flexibility of our proposed framework.

\section{Experiment}
\label{experiments}

\subsection{Experimental Setup}
\subsubsection{Backbone Networks and Datasets}
In C3DS, we evaluate the proposed SeCondPoint framework with $3$ typical 3D point cloud segmentation backbone networks: DGCNN~\cite{2018dgcnn}, RandLA-Net~\cite{2020randlanet}, and SCF-Net~\cite{fan2021scf}. In this work, the architectures of the used networks are totally the same with those in their original papers since we directly use their public codes and models. We choose the three networks not only because their codes and models are public, but also because they are representative networks. Specifically, DGCNN is a graph-based architecture and usually used to process block-size small-scale point clouds. While RandLA-Net and SCF-Net are two different point-based networks, which can be directly used to deal with large-scale point clouds. Two public 3D point cloud scene datasets, S3DIS~\cite{2016s3dis} and ScanNet~\cite{2017scannet}, are used to evaluate the proposed SeCondPoint framework. The details about the two datasets are provided in the supplemental materials due to the limited space. The semantic feature of each given class is determined simply by 1) recording its class name and 2) retrieving its semantic feature among all word2vec embeddings~\cite{MikolovSCCD13} according to its name, which is represented by a $300$-D vector. 

In Z3DS, we evaluate the proposed SeCondPoint framework with DGCNN and RandLA-Net being the network backbones. Other backbones could be adapted seamlessly to Z3DS in the same way. We construct two benchmarks by re-organizing S3DIS~\cite{2016s3dis} and ScanNet~\cite{2017scannet}. In S3DIS, we consider the $12$ semantic categories as valid categories and ignore the `clutter' class, which are split into seen classes and unseen classes in $3$ different manners, i.e. 10/2, 8/4, 6/6 as seen classes/unseen classes respectively. In ScanNet, we choose $19$ semantic categories as valid categories, which are also split in $3$ different ways, i.e. 16/3, 13/6, 10/9 splits. More details are introduced in the supplemental materials due to the limited space. For the sake of clarity, we denote the re-organized S3DIS and ScanNet as S3DIS-0 and ScanNet-0 respectively. The semantic features are the same with those in C3DS.

\begin{table*}[t]
	\begin{center}
		\caption{Comparative results (\%) on S3DIS (Area-5).} \label{tab2}
		\resizebox{2.0\columnwidth}{!}{
			\begin{tabular}{c|c|c|c|c|c|c|c|c|c|c|c|c|c|c|c|c}
				\hline
				\textbf{Method} &\textbf{$mACC$}	&\textbf{$mIoU$}	&\textbf{$OA$}	&ceiling	&floor	&wall	&beam	&column	&window	&door	&table	&chair	&sofa	&bookcase	&board	&clutter\\
				\hline
				SPGraph~\cite{2018spgraph}	        &66.5   &58.0 	&86.4	&89.4 	&96.9 	&78.1 	&0.0 	&42.8 	&48.9 	&61.6 	&84.7 	&75.4 	&69.8 	&52.6 	&2.1 	&52.2\\
				PointCNN~\cite{2018pointcnn}	    &63.9   &57.3 	&86.0	&92.3 	&98.2 	&79.4 	&0.0 	&17.6 	&22.8 	&62.1 	&74.4 	&80.6 	&31.7 	&66.7 	&62.1 	&56.7\\
				PointWeb~\cite{2019pointweb}	    &66.6   &60.3 	&87.0	&92.0 	&98.5 	&79.4 	&0.0 	&21.1 	&59.7 	&34.8 	&76.3 	&88.3 	&46.9 	&69.3 	&64.9 	&52.5\\
				Point2Node~\cite{2020point2node}	&70.0   &63.0 	&\textbf{88.8}	&93.9 	&98.3 	&83.3 	&0.0 	&35.7 	&55.3 	&58.8 	&79.5 	&84.7 	&44.1 	&71.1 	&58.7 	&55.2\\
				\hline
				\hline
				DGCNN~\cite{2018dgcnn} 				    &56.3   &49.8 	&85.7	&91.3 	&97.4 	&77.9 	&0.0 	&9.1 	&52.4 	&20.3 	&70.3 	&72.0 	&24.6 	&50.7 	&35.8 	&45.8\\
				DGCNN+RSMix~\cite{lee2021regularization}&55.2{\color{blue}(-1.1)}   &48.1{\color{blue}(-1.7)} 	&84.2{\color{blue}(-1.5)}	&87.5   &92.7   &74.2   &0.0    &9.0    &41.8   &11.8   &67.7   &69.0   &26.6   &56.1   &44.3   &45.2\\
				DGCNN+SeCondPoint         			    &62.8{\color{blue}(+6.5)}   &52.7{\color{blue}(+2.9)} 	&86.0{\color{blue}(+0.3)}	&91.0 	&97.3 	&77.4 	&0.0 	&16.2 	&56.8 	&31.1 	&68.6 	&69.0 	&32.3 	&53.4 	&44.6 	&47.4\\
				\hline
				RandLA-Net~\cite{2020randlanet}	                &70.3   &62.5 	&87.0	&92.1 	&96.9 	&80.0 	&0.0 	&24.8 	&60.9 	&35.5 	&77.2 	&85.0 	&72.6 	&70.3 	&67.1 	&50.6\\ 
				RandLA-Net+RSMix~\cite{lee2021regularization}   &70.8{\color{blue}(+0.5)}   &63.1{\color{blue}(+0.6)} 	&87.5{\color{blue}(+0.5)}	&93.0   &97.4   &80.6   &0.0    &23.1   &58.9   &39.9   &78.1   &86.6   &67.5   &71.3   &71.1   &53.2\\
				RandLA-Net+SeCondPoint                          &\textbf{74.6}{\color{blue}(+4.3)}   &64.2{\color{blue}(+1.7)} 	&87.6{\color{blue}(+0.6)}	&92.2 	&97.3 	&80.7 	&0.0 	&34.5 	&61.8 	&47.1 	&76.4 	&84.1 	&75.6 	&71.2 	&63.0 	&51.2\\
				\hline
				SCF-Net~\cite{fan2021scf}	       			 &71.5   &63.5 	&87.2	&90.8   &96.9   &81.0   &0.0    &21.2   &60.3   &44.2   &78.9   &87.7   &74.1   &71.4   &68.6   &50.3\\ 
				SCF-Net+RSMix~\cite{lee2021regularization}	 &70.8{\color{blue}(-0.7)}   &62.2{\color{blue}(-1.3)} 	&86.4{\color{blue}(-0.8)}	&89.6   &95.1   &79.9   &0.0    &22.0   &62.3   &34.3   &76.7   &87.8   &70.2   &70.7   &70.7   &49.9\\ 
				SCF-Net+SeCondPoint                 		 &74.2{\color{blue}(+2.7)}   &\textbf{64.4}{\color{blue}(+0.9)} 	&87.6{\color{blue}(+0.4)}	&89.9  &97.1  &80.9   &0.0  &23.8  &61.8  &48.0  &78.4  &87.3  &76.8  &72.1  &71.3  &49.8\\
				\hline
			\end{tabular}
		}
	\end{center}
\end{table*}

\begin{table*}[t]
	\begin{center}
		\caption{Comparative results (\%) on S3DIS (6-fold).} \label{tab3}
		\resizebox{2.0\columnwidth}{!}{
			\begin{tabular}{c|c|c|c|c|c|c|c|c|c|c|c|c|c|c|c|c}
				\hline
				\textbf{Method}       &\textbf{$mACC$}  &\textbf{$mIoU$}	&\textbf{$OA$}	&ceiling	&floor	&wall	&beam	&column	&window	&door	&table	&chair	&sofa	&bookcase	&board	&clutter\\
				\hline
				SPGraph~\cite{2018spgraph}	 	   &73.0   &62.1	&85.5	&89.9	&95.1	&76.4	&62.8	&47.1	&55.3	&68.4	&73.5	&69.2	&63.2	&45.9	&8.7	&52.9\\
				PointCNN~\cite{2018pointcnn}	   &75.6   &65.4	&88.1	&94.8	&97.3	&75.8	&63.3	&51.7	&58.4	&57.2	&71.6	&69.1	&39.1	&61.2	&52.2	&58.6\\
				PointWeb~\cite{2019pointweb}	   &76.2   &66.7	&87.3	&93.5	&94.2	&80.8	&52.4	&41.3	&64.9	&68.1	&71.4	&67.1	&50.3	&62.7	&62.2	&58.5\\
				Point2Node~\cite{2020point2node}   &79.1   &70.0	&89.0	&94.1	&97.2	&83.4	&62.7	&52.3	&72.3	&64.3	&75.8	&70.8	&65.7	&49.8	&60.3	&60.9\\
				\hline
				\hline
				DGCNN~\cite{2018dgcnn}					&68.5   &57.2 	&85.8	&93.2 	&94.8 	&76.6 	&37.3 	&33.1 	&50.7 	&62.3 	&62.4 	&63.4 	&25.2 	&47.0 	&45.6 	&52.9\\
				DGCNN+RSMix~\cite{lee2021regularization}&69.4{\color{blue}(+0.9)}   &58.8{\color{blue}(+1.6)} 	&85.0{\color{blue}(-0.8)}	&86.7   &89.7   &76.4   &56.4   &34.9   &51.3   &60.1   &63.4   &65.1   &28.8   &49.7   &50.2   &52.2\\
				DGCNN+SeCondPoint         				&72.9{\color{blue}(+4.4)}   &58.4{\color{blue}(+1.2)} 	&86.5{\color{blue}(+0.7)}	&93.3 	&96.4 	&75.1 	&38.3 	&35.5 	&49.6 	&66.4 	&62.1 	&68.1 	&29.5 	&47.4 	&44.7 	&52.9\\
				\hline
				RandLA-Net~\cite{2020randlanet}	    		  &79.6   &68.7 	&88.1	&94.0 	&96.6 	&80.8 	&43.4 	&45.7 	&62.7 	&70.8 	&66.9 	&78.6 	&64.7 	&61.4 	&66.0 	&61.4\\
				RandLA-Net+RSMix~\cite{lee2021regularization} &81.0{\color{blue}(+1.4)}   &68.9{\color{blue}(+0.2)} 	&87.5{\color{blue}(-0.6)}	&93.4   &96.8   &79.3   &54.8   &43.9   &62.5   &67.2   &70.2   &76.9   &60.3   &64.1   &66.1   &59.6\\
				RandLA-Net+SeCondPoint	    				  &82.1{\color{blue}(+2.5)}   &69.5{\color{blue}(+0.8)} 	&89.0{\color{blue}(+0.9)}	&94.6 	&97.3 	&81.9 	&44.3 	&48.5 	&62.3 	&72.4 	&66.5 	&80.8 	&63.6 	&63.6 	&65.6 	&62.3\\
				\hline
				SCF-Net~\cite{fan2021scf}	    			&81.9   &70.2 	&88.8	&94.0  &96.5  &81.0  &48.6  &47.8  &62.0  &72.9  &68.4  &82.4  &67.3  &62.8  &65.6  &63.8\\
				SCF-Net+RSMix~\cite{lee2021regularization}  &82.1{\color{blue}(+0.2)}   &70.7{\color{blue}(+0.5)} 	&88.0{\color{blue}(-0.8)}	&92.6  &95.7  &80.2  &62.8  &46.6  &65.2  &68.2  &70.2  &81.4  &66.4  &64.9  &64.8  &60.3\\
				SCF-Net+SeCondPoint	    					&\textbf{82.9}{\color{blue}(+1.0)}   &\textbf{71.4}{\color{blue}(+1.2)} 	&\textbf{89.5}{\color{blue}(+0.7)}	&93.8  &96.4  &81.4  &53.2  &50.9  &65.8  &73.1  &68.5  &82.8  &67.7  &63.2  &67.2  &63.7\\
				\hline
			\end{tabular}
		}
	\end{center}
\end{table*}

\subsubsection{Evaluation Protocols and Comparative Methods} 
For C3DS, following most existing works~\cite{2018spgraph,2018pointcnn,2018dgcnn,2020randlanet,fan2021scf}, we use overall Top-1 accuracy ($OA$), average per-class Top-1 accuracy ($mACC$), and average per-class IoU ($mIoU$) to evaluate 3D semantic segmentation performance on S3DIS. The models are tested in two different settings for a comprehensive evaluation, i.e. 1) testing on Area $5$ and training on the remaining $5$ areas (in short, Area-$5$ validation), and 2) $6$-fold cross validation. Semantic segmentation on ScanNet is usually regarded as a voxel-level semantic labeling task, hence overall Top-1 voxel accuracy ($OVA$) and average per-class Top-1 voxel accuracy ($mVACC$) are used for evaluation as done in~\cite{2020point2node, 2019pointweb}.

For Z3DS, according to the evaluation protocols in zero-shot learning~\cite{Xian17Comprehensive} and 3D semantic segmentation, we use average per-class Top-1 accuracy and average per-class IoU on unseen classes to evaluate the Z3DS performance on S3DIS-0, denoted by ($mACC_{u}$) and ($mIoU_{u}$) respectively, and only the Area-$5$ validation are used considering that Area-$5$ validation can better test the model's generalization ability since 3D scenes in Area-$5$ have no overlap with those in other areas. On ScanNet-0, average per-class Top-1 accuracy ($mACC_{u}$) and average per-class voxel accuracy ($mVACC_{u}$) on unseen classes are used. In GZ3DS, as done in the generalized zero-shot learning~\cite{Xian17Comprehensive}, we first compute $mACC_{u}$ (also $mIoU_{u}$ and $mVACC_{u}$) and $mACC_{s}$ (also $mIoU_{s}$ and $mVACC_{s}$) on unseen classes and seen classes respectively, and then their harmonic mean $HACC$ (also $HIoU$ and $HVACC$) is computed to evaluate the overall performance by:
\begin{equation}
	HACC = \frac{2 \times mACC_{s} \times mACC_{u}}{mACC_{s} + mACC_{u}}
	\label{eq6}
\end{equation}
Here, we only describe the computation of $HACC$ in (\ref{eq6}), while $HIoU$ and $HVACC$ are computed in the same way. On S3DIS-0, the overall performance is evaluated by $HACC$ and $HIoU$, while on ScanNet-0, it is evaluated by $HACC$ and $HVACC$. Note that the reason for using the harmonic mean is that many GZSL methods~\cite{liu2020zero, liu2021iterative, liu2021hardness} generally suffer from the bias towards seen classes, in other words, their performances on seen classes are significantly higher than those on unseen classes, hence the harmonic mean is used here as done in these methods to emphasize the importance of unseen-class data classification.

In C3DS, the proposed methods are compared with $10$ existing state-of-the-art networks, including DGCNN~\cite{2018dgcnn}, RandLA-Net~\cite{2020randlanet}, SCF-Net~\cite{fan2021scf}, PointNet++~\cite{2017pointnet++}, SPGraph~\cite{2018spgraph}, PointCNN~\cite{2018pointcnn}, ACNN~\cite{2019acnn}, PointWeb~\cite{2019pointweb}, Point2Node~\cite{2020point2node}, and PointGCR~\cite{2020pointgcr}. Besides, we also adapt a data augmentation approach originally for 3D shape classification to deal with 3D scene semantic segmentation, and compare our proposed framework with the adapted method. In Z3DS, since this work is the first attempt to investigate the Z3DS task, we firstly establish two baseline Z3DS methods by straightforwardly introducing two zero-shot learning methods~\cite{Frome13DeViSE,Xian18FCLSWGAN} into Z3DS, and then we compare the proposed framework with the two baselines.

\subsubsection{Implementation Details}
The used three backbone architectures in this paper are totally consistent with the corresponding original backbone networks. In C3DS, we directly use the public pre-trained models provided by the authors or train the models with the public codes according to the hyper-parameters given by the authors. In Z3DS, the backbone networks are trained from scratch with the seen-class objects using the public codes and hyper-parameters. In practice, a whole point cloud is input into the models, however, only the labeled seen-class points are used to compute gradients and the unlabeled unseen-class points are not computed in the back-propagation process. We consider that this training setting satisfies the standard of zero-shot learning and it is not a transductive setting because 1) the unlabeled unseen-class points do not provide any form of supervision signal for model learning and 2) the testing point clouds are not used at the training stage. Actually, in the zero-shot 2D semantic segmentation works~\cite{bucher2019zero,GuZ00Z20,XianC0SA19}, unlabeled unseen-class pixels are also input into the model at the training stage for not destroying the image structure. 

The feature generator, discriminator and classifier are multi-layer fully-connected neural networks, and the corresponding architectures are detailed in the supplemental materials. The generator and discriminator are adversarially trained by $20$ epochs, with a batch size of $32$ and a learning rate of $0.0005$. The point feature classifier is trained by $10$ epochs, with a batch size of $4096$ and a learning rate of $0.0001$. All models are learned with the Adam optimizer.

\begin{table}[t]
	\begin{center}
		\caption{Comparative results (\%) on ScanNet.} \label{tab4}
		\resizebox{0.65\columnwidth}{!}{
			\begin{tabular}{c|c|c}
				\hline
				\textbf{Method}	        &\textbf{$mVACC$}  &\textbf{$OVA$}\\
				\hline
				PointNet++~\cite{2017pointnet++}   &-    	&84.5\\
				PointCNN~\cite{2018pointcnn}	   &-    	&85.1\\
				ACNN~\cite{2019acnn}	           &-    	&85.4\\
				PointWeb~\cite{2019pointweb}	   &60.6    &85.9\\
				PointGCR~\cite{2020pointgcr}	   &60.5 	&85.3\\
				\hline
				\hline
				RandLA-Net~\cite{2020randlanet}	   				   &59.5     &85.7 \\
				RandLA-Net+RSMix~\cite{lee2021regularization}	   &60.3{\color{blue}(+0.8)}     &85.6{\color{blue}(-0.1)} \\
				RandLA-Net+SeCondPoint	   		   				   &65.6{\color{blue}(+6.1)}     &85.9{\color{blue}(+0.2)}  \\
				\hline
				SCF-Net~\cite{fan2021scf}	   				   &61.4     &\textbf{86.0}\\
				SCF-Net+RSMix~\cite{lee2021regularization}	   &61.5{\color{blue}(+0.1)}     &85.9{\color{blue}(-0.1)} \\
				SCF-Net+SeCondPoint	   		   				   &\textbf{68.9}{\color{blue}(+7.5)}     &85.9{\color{blue}(-0.1)}\\
				\hline
			\end{tabular}
		}
	\end{center}
\end{table}

\subsection{C3DS Results on S3DIS}
Here we evaluate the proposed SeCondPoint by embedding DGCNN~\cite{2018dgcnn}, RandLA-Net~\cite{2020randlanet}, and SCF-Net~\cite{fan2021scf} into the proposed framework on S3DIS under both the Area-$5$ validation and the $6$-fold cross validation settings. Since the pre-trained models of DGCNN are not released by the authors, we firstly pre-train DGCNN with the public codes and the corresponding hyper-parameters provided by the authors. For RandLA-Net and SCF-Net, we directly use the pre-trained models released by the authors\footnote{https://github.com/QingyongHu/RandLA-Net, https://github.com/leofansq/SCF-Net}. The results of these pre-trained models and the corresponding enhanced models by the proposed framework in the Area-$5$ validation and the $6$-fold cross validation are reported in Table~\ref{tab2} and Table~\ref{tab3} respectively. We also adapt a data augmentation method (RSMix~\cite{lee2021regularization}) originally proposed for 3D shape classification to handle the C3DS task, and the results of DGCNN, RandLA-Net, and SCF-Net augmented by RSMix are reported in the two tables too. In addition, we report the results of some state-of-the-art 3D segmentation methods as shown in Table~\ref{tab2} and Table~\ref{tab3} for further comparison.

As noted from Table~\ref{tab2} and Table~\ref{tab3}, five points are revealed. Firstly, the performances of all the three backbone networks are significantly improved by the proposed framework in both the used validations in terms of $mACC$, $mIoU$, and $OA$. For instance, the $mACC$ of DGCNN is improved by $6.5\%$ and $4.4\%$ in the Area-$5$ validation and the $6$-fold cross validation respectively, indicating that the proposed framework is able to generate many augmented point features from the semantics conditioned feature distributions for learning a better point classifier. 

Secondly, compared with the state-of-the-art data augmentation method (RSMix~\cite{lee2021regularization}) from the perspective of data augmentation, the proposed framework improves all the backbone networks by larger margins in both the used validations. In fact, it is noticed that RSMix even decreases the performances of the backbone networks at some cases. This is probably because RSMix is originally designed to augment single 3D object shapes for 3D shape classification, and it would hurt the structures of 3D scenes when applied to 3D scene segmentation. 

Thirdly, the improvements of $mACC$ are more significant than those of $mIoU$ and $OA$ under the proposed framework for all the three backbone networks. This is mainly because for those classes with a relatively small number of points in the original training point clouds, the diversity of their point features are largely augmented by the proposed feature modeling network, and consequently their accuracies are improved. This is consistent with the results shown in Table~\ref{tab2} and Table~\ref{tab3} where some classes (e.g. sofa, column) that originally have low performances are significantly improved by the proposed framework. Both support the explanation in the caption of Figure~\ref{fig1}. 

Fourthly, the improvements in the Area-$5$ validation are more significant than those in the $6$-fold cross validation for the three backbone networks in most cases. Considering that the validation on Area $5$ is harder than that on the other areas (since 3D scenes in Area $5$ have no overlap with those in other areas), the larger improvements in the Area-$5$ validation further demonstrate the generalization ability of the proposed framework. 

Finally, the enhanced SCF-Net by the proposed framework outperforms the state-of-the-art methods significantly in most cases, especially in terms of $mACC$, by margins about $4.2\%$ and $3.8\%$ in the Area-$5$ validation and the $6$-fold cross validation respectively.

\subsection{C3DS Results on ScanNet}
We also validate the proposed SeCondPoint framework on ScanNet with RandLA-Net and SCF-Net. We do not conduct experiments with DGCNN on ScanNet because it is needed to split a whole point cloud into blocks and randomly sample points in blocks when applying DGCNN on Scannet, but a unified splitting protocol does not exist in literature, hence a fair comparison with existing methods could hardly be made. For both RandLA-Net and SCF-Net, the pre-processing operations (only down sampling) are simple and unified. Since the results of RandLA-Net and SCF-Net on ScanNet are not directly reported by their authors, we firstly obtain their results by training them according to the public codes and the hyper-parameters provided by the authors, which are reported in Table~\ref{tab4}. Then, we enhance RandLA-Net and SCF-Net with the proposed SeCondPoint framework and a state-of-the-art data augmentation technique (i.e. RSMix~\cite{lee2021regularization}) respectively, and the corresponding results are reported in Table~\ref{tab4}. We also report the results of several state-of-the-art methods in Table~\ref{tab4} for further comparison. As seen from Table~\ref{tab4}, firstly, the $OVA$ of the enhanced RandLA-Net and SCF-Net by the proposed framework are close to those of the original methods, but the $mVACC$ of the two enhanced methods are significantly improved by about $6.1\%$ and $7.5\%$ respectively. This is because those classes with a relatively small number of points originally are augmented by the proposed framework, and the significant increases of the accuracies of those classes result in a better average per-class accuracy. Secondly, the improvements achieved by the proposed framework is significantly better than those achieved by RSMix. Actually, RSMix makes a negligible effect on both RandLA-Net and SCF-Net in ScanNet, which is probably caused by the fact that it is designed to augment 3D object shapes for 3D shape classification. Finally, both the enhanced SCF-Net and the enhanced RandLA-Net outperform all the comparative state-of-the-art methods by large margins, demonstrating the effectiveness of the proposed framework.

\begin{table*}
	\centering
	\resizebox{1.6\columnwidth}{!}{
		\begin{tabular}{cc|cc|cccccc}
			\hline
			\multicolumn{2}{c|}{Method}  & \multicolumn{2}{c|}{Z3DS} & \multicolumn{6}{c}{GZ3DS} \\
			\hline
			&  &$mACC_{u}$  &$mIoU_{u}$  &$HACC$  &$mACC_{s}$  &$mACC_{u}$  &$HIoU$  &$mIoU_{s}$  &$mIoU_{u}$  \\
			\hline
			\multirow{8}{*}{10/2}
			&DGCNN+Fully-Supervised   	&-  &-    &64.7 &58.1  &72.9 &58.8 &52.4  &67.0  \\
			&DGCNN+3D-V2S-embedding &92.1  &86.1  &0.0  &52.2  &0.0  &0.0  &39.1  &0.0  \\
			&DGCNN+3D-GAN      	&93.6  &88.7  &55.5  &51.5  &60.3  &29.9  &44.1  &22.7  \\
			&DGCNN-Z3DS(ours)       &95.8  &92.5  &56.8  &49.2  &67.0  &31.0  &43.1  &24.2  \\
			\cline{2-10}
			&RandLA-Net+Fully-Supervised &-  &-  	   &75.3  &68.8  &83.2  &67.8  &63.3  &73.1  \\
			&RandLA-Net+3D-V2S-embedding &93.5  &87.8  &0.0   &70.3  &0.0   &0.0  &57.3  &0.0  \\
			&RandLA-Net+3D-GAN      	 &93.9  &88.6  &55.3  &55.3  &55.3  &31.0  &42.2  &24.5  \\
			&RandLA-Net-Z3DS(ours)  	 &95.6  &91.6  &60.0  &51.5  &71.9  &31.4  &40.7  &25.6  \\
			\hline
			\multirow{8}{*}{8/4}
			&DGCNN+Fully-Supervised   	&-  &-        &58.0  &64.5  &52.7  &52.7  &58.2  &48.2  \\
			&DGCNN+3D-V2S-embedding &46.5  &18.4  &0.0   &67.8  &0.0   &0.0   &52.1  &0.0  \\
			&DGCNN+3D-GAN      	&46.3  &18.1  &46.0  &57.4  &38.4  &9.6   &51.8  &5.3  \\
			&DGCNN-Z3DS(ours)       &48.9  &19.7  &45.4  &58.4  &37.1  &10.2  &51.6  &5.6  \\
			\cline{2-10}
			&RandLA-Net+Fully-Supervised &-  &-        &71.4  &70.7  &72.2  &64.6  &65.6  &63.6  \\
			&RandLA-Net+3D-V2S-embedding &46.0  &13.7  &0.2   &71.6  &0.1   &0.2   &59.5  &0.1  \\
			&RandLA-Net+3D-GAN      	 &52.1  &23.9  &39.2  &37.5  &41.0  &9.4   &29.6  &5.6  \\
			&RandLA-Net-Z3DS(ours)  	 &55.3  &26.0  &41.9  &60.0  &32.1  &14.1  &49.4  &8.2  \\
			\hline
			\multirow{8}{*}{6/6}
			&DGCNN+Fully-Supervised   	&-  &- 		  &56.2  &76.8 &44.3 &51.1 &69.2  &40.5  \\
			&DGCNN+3D-V2S-embedding &30.3  &11.3  &0.0  &76.6  &0.0  &0.0  &60.0  &0.0  \\
			&DGCNN+3D-GAN      	&30.3  &13.7  &33.2 &52.0  &24.4 &5.6  &48.7  &3.0  \\
			&DGCNN-Z3DS(ours)       &32.2  &14.3  &32.1 &46.6  &24.4 &6.0  &44.4  &3.2  \\
			\cline{2-10}
			&RandLA-Net+Fully-Supervised &-  &-  	   &70.3  &79.0  &63.3  &63.7  &73.8  &56.0  \\
			&RandLA-Net+3D-V2S-embedding &33.0  &8.9   &1.4   &79.5  &0.7   &1.1   &64.7  &0.6  \\
			&RandLA-Net+3D-GAN      	 &32.8  &19.6  &33.5  &42.3  &27.8  &4.8   &39.9  &2.6  \\
			&RandLA-Net-Z3DS(ours)  	 &33.9  &20.8  &34.8  &36.5  &33.2  &6.9   &36.2  &3.8  \\
			\hline
		\end{tabular}
	}
	\caption{Comparative results on the S3DIS-0 dataset.}
	\label{tab5}
\end{table*}

\begin{table*}
	\centering
	\resizebox{1.6\columnwidth}{!}{
		\begin{tabular}{cc|cc|cccccc}
			\hline
			\multicolumn{2}{c|}{Method}  & \multicolumn{2}{c|}{Z3DS} & \multicolumn{6}{c}{GZ3DS} \\
			\hline
			&  &$mACC_{u}$  &$mVACC_{u}$  &$HACC$  &$mACC_{s}$  &$mACC_{u}$  &$HVACC$  &$mVACC_{s}$  &$mVACC_{u}$  \\
			\hline
			\multirow{8}{*}{16/3}
			&DGCNN+Fully-Supervised   	&-  &-     &39.1  &42.3  &36.5  &43.2  &45.0  &41.6  \\
			&DGCNN+3D-V2S-embedding  &57.7  &58.7  &0.0  &36.9  &0.0  &0.0  &40.1  &0.0  \\
			&DGCNN+3D-GAN      	     &60.4  &63.5  &24.7 &43.4  &17.3 &26.4 &46.9  &18.4  \\
			&DGCNN-Z3DS(ours)        &63.5  &65.4  &29.1 &40.9  &22.5 &31.1 &43.8  &24.2  \\
			\cline{2-10}
			&RandLA-Net+Fully-Supervised   	&-  &-     &57.5 &60.9  &54.4 &58.1 &61.5  &55.1  \\
			&RandLA-Net+3D-V2S-embedding &55.3  &54.7  &0.0  &36.7  &0.0  &0.0  &37.2  &0.0  \\
			&RandLA-Net+3D-GAN      	 &60.5  &62.1  &40.9 &58.0  &31.6 &41.0 &58.3  &31.6  \\
			&RandLA-Net-Z3DS(ours)  	 &63.9  &65.7  &42.1 &58.8  &32.8 &42.2 &59.0  &32.9  \\
			\hline
			\multirow{8}{*}{13/6}
			&DGCNN+Fully-Supervised   	&-  &-    &40.4 &42.7  &38.4 &43.8 &45.5  &42.2  \\
			&DGCNN+3D-V2S-embedding &21.9  &23.7  &0.0  &47.2  &0.0  &0.0  &51.0  &0.0  \\
			&DGCNN+3D-GAN      	    &43.9  &45.0  &28.8 &29.7  &27.9 &30.2 &31.1  &29.3  \\
			&DGCNN-Z3DS(ours)       &45.7  &47.0  &29.2 &29.1  &29.3 &30.8 &33.0  &28.8  \\
			\cline{2-10}
			&RandLA-Net+Fully-Supervised   	&-  &-     &58.4 &62.1  &55.1 &59.3 &62.3  &56.6  \\
			&RandLA-Net+3D-V2S-embedding &22.5  &22.5  &0.2  &62.5  &0.1  &0.1  &62.6  &0.1  \\
			&RandLA-Net+3D-GAN      	 &41.1  &42.4  &30.0  &48.8  &21.7  &30.8  &49.0  &22.4  \\
			&RandLA-Net-Z3DS(ours)  	 &45.1  &46.5  &33.1  &35.7  &30.8  &35.2  &37.9  &32.9  \\
			\hline
			\multirow{8}{*}{10/9}
			&DGCNN+Fully-Supervised   	&-  &-    &41.4  &40.4  &42.4  &44.5  &42.3  &46.9  \\
			&DGCNN+3D-V2S-embedding &20.8  &21.8  &0.0  &49.4  &0.0  &0.0  &49.4  &0.0  \\
			&DGCNN+3D-GAN      	    &26.8  &27.2  &20.6 &34.7  &14.7 &20.8 &37.7  &14.4  \\
			&DGCNN-Z3DS(ours)       &27.4  &27.5  &22.0 &34.8  &16.1 &22.6 &38.0  &16.1  \\
			\cline{2-10}
			&RandLA-Net+Fully-Supervised   	&-  &-     &59.9 &59.1  &60.7 &60.5 &59.2  &61.8  \\
			&RandLA-Net+3D-V2S-embedding &21.8  &22.1  &1.1  &26.1  &0.6  &1.1  &26.2  &0.6  \\
			&RandLA-Net+3D-GAN      	 &29.2  &29.5  &20.6 &60.8  &12.4 &20.8 &62.4  &12.5  \\
			&RandLA-Net-Z3DS(ours)  	 &30.9  &31.2  &23.4 &59.1  &14.6 &24.0 &60.4  &15.0  \\
			\hline
		\end{tabular}
	}
	\caption{Comparative results on the ScanNet-0 dataset.}
	\label{tab6}
\end{table*}

\subsection{Z3DS/GZ3DS Results on S3DIS-0}
To demonstrate the Z3DS/GZ3DS ability of the proposed framework, here we evaluate the proposed framework in both Z3DS and GZ3DS by embedding existing segmentation networks (i.e. DGCNN~\cite{2018dgcnn} and RandLA-Net~\cite{2020randlanet}) into the proposed framework on S3DIS-0 with $3$ different seen/unseen splits in the Area-$5$ validation. The enhanced DGCNN and RandLA-Net are denoted by DGCNN-Z3DS and RandLA-Net-Z3DS respectively for easy reading. We also implement two baselines by straightforwardly embedding DGCNN and RandLA-Net into two typical zero-shot learning methods~\cite{Frome13DeViSE,Xian18FCLSWGAN}. \cite{Frome13DeViSE} is a visual-to-semantic embedding based zero-shot learning method while \cite{Xian18FCLSWGAN} is a GAN based method. For the sake of clarity, we denote them by 3D-V2S-embedding and 3D-GAN respectively. The results of our proposed methods and the two baselines are all reported in Table~\ref{tab5}. Besides, the results of the fully-supervised model (the labels of seen classes and unseen classes are given in training) in GZ3DS are reported so that the performance gap between the zero-shot segmentation methods and the corresponding fully-supervised ones can be seen clearly.

Five observations can be made from Table~\ref{tab5}. Firstly, in Z3DS, the accuracies of both the proposed framework and the baselines exceed $90\%$ in the 10/2 split. This demonstrates that novel-class inference could be realistic in the 3D scene segmentation task by introducing the language-level semantics at least when the number of novel classes is not very large.

Secondly, $HACC$ and $HIoU$ of 3D-V2S-embedding in GZ3DS are both $0$, demonstrating that 3D-V2S-embedding severely suffers from the bias towards seen classes. This problem is partly caused by the facts that the feature learning of seen-class objects and unseen-class objects in 3D scenes are mutually affected, and since the unseen-class objects have no supervised signal in the feature learning process, their features easily bias towards those of seen classes. Note that this phenomenon is significantly different from that in generalized zero-shot 3D shape classification~\cite{cheraghian2020transductive} where a relatively better result is achieved, showing the difference between the two tasks. 

Thirdly, the proposed framework achieves significantly superior performances over the two baseline methods in all data splitting cases in both Z3DS and GZ3DS, especially when compared with 3D-V2S-embedding. These large improvements demonstrate the effectiveness of the proposed semantics conditioned feature modeling and the feature-geometry-based mixup training approach. 

Fourthly, compared with the fully-supervised model, the GZ3DS performances of the proposed framework in the 10/2 split are relatively lower but still encouraging, considering the fact that the unseen-class objects totally have no supervised signal in the training under the proposed framework. These results demonstrate that novel classes in realistic 3D scenes can be well handled at least when the number of novel classes is not very large. 

Finally, the Z3DS and GZ3DS performances of the proposed framework (also 3D-V2S-embedding and 3D-GAN) decrease significantly as the number of unseen classes increases. This indicates that it is still a hard problem to infer a large number of unseen classes with a limited number of seen classes, and a considerable number of seen classes are helpful for achieving high-performance Z3DS/GZ3DS. We leave this direction as a future work.

\subsection{Z3DS/GZ3DS Results on ScanNet-0}
Here we validate the proposed framework on ScanNet-0 with $3$ different seen/unseen splits in both Z3DS and GZ3DS. The employed backbone networks (i.e. DGCNN and RandLA-Net) and the two comparative baselines (i.e. 3D-V2S-embedding and 3D-GAN) are the same with those in the S3DIS-0 benchmark. All the results are reported in Table~\ref{tab6}. As seen from Table~\ref{tab6}, the Z3DS performances of the proposed framework are significantly better than those of the two baselines, for instance, $mACC_{u}$ of DGCNN+3D-V2S-embedding and DGCNN+3D-GAN in the 16/3 split are $57.7\%$ and $60.4\%$ respectively, while that of DGCNN-Z3DS reach $63.5\%$. Secondly, in GZ3DS, 3D-V2S-embedding severely suffers from the bias problem like in S3DIS-0, while the proposed framework outperforms the baselines with large margins, for instance, the $HVACC$ of RandLA-Net-Z3DS in the 16/3 split reaches $42.2\%$ while that of RandLA-Net+3D-V2S-embedding is $0$, which demonstrates that the proposed semantics conditioned feature modeling is helpful to alleviate the bias problem. Besides, the improvements achieved by the proposed framework over 3D-GAN further demonstrate the effectiveness of the proposed framework.

\subsection{Result Analysis}
\subsubsection{Effect of Feature-Geometry-Based Mixup}
\begin{figure}
	\centering
	\includegraphics[width=0.9\linewidth]{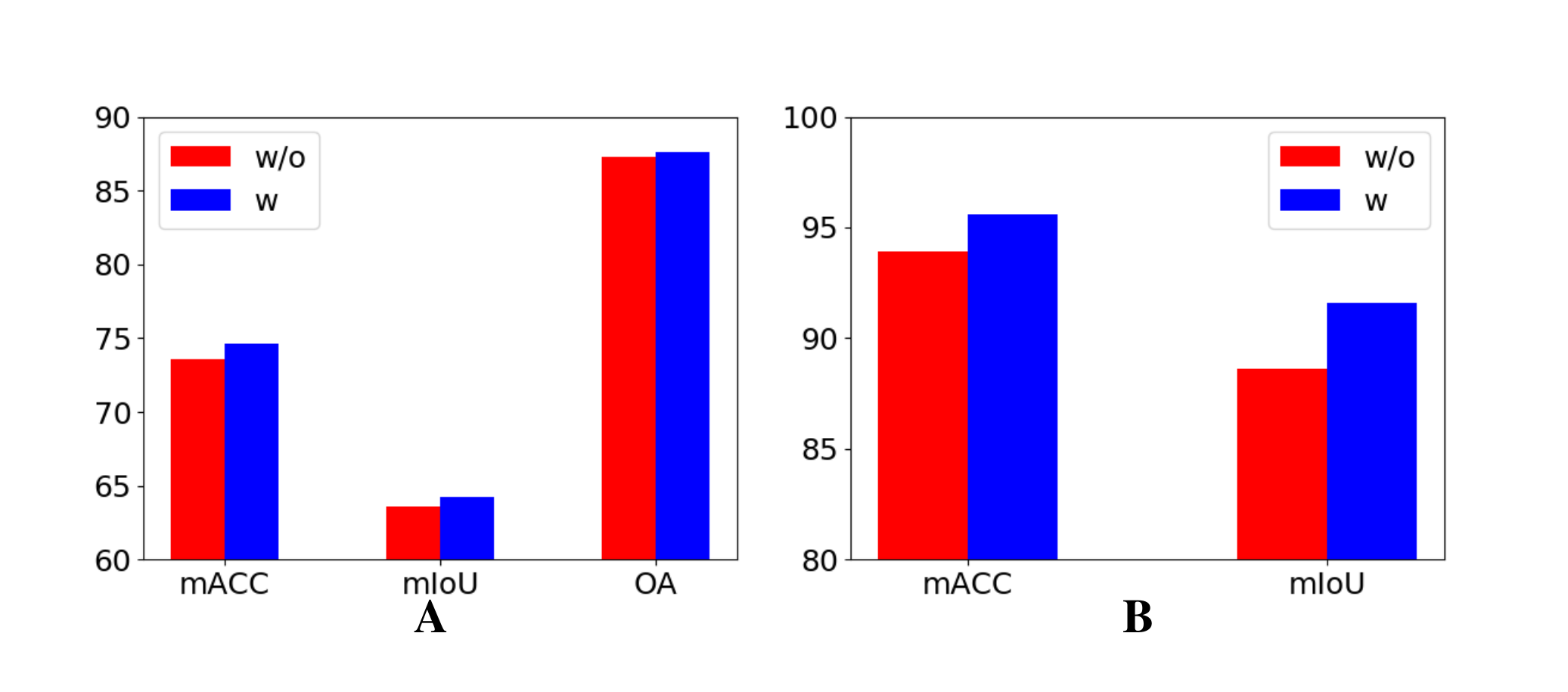}
	\caption{Comparative results (\%) with/without feature-geometry-based mixup training. A for C3DS and B for Z3DS.}
	\label{fig3}
\end{figure}

Here we analyze the effect of the feature-geometry-based mixup approach. We conduct experiments in both the C3DS task and the Z3DS task using RandLA-Net as backbone network in the Area-$5$ validation setting on S3DIS. The results of C3DS and Z3DS are shown in Figure~\ref{fig3}-A and Figure~\ref{fig3}-B respectively, where `w' means using feature-geometry-based mixup training while `w/o' means not using it. Note that the Z3DS task is performed under the 10/2 split. As seen from Figure~\ref{fig3}, the feature-geometry-based mixup training clearly improves the model's performances in terms of $mACC$, $mIoU$, and $OA$ in C3DS and Z3DS. This is mainly because by interpolating between two adjacent classes in the feature space, the local discrimination between classes could be enhanced.

\subsubsection{Effect of Number of Generated Features}
\begin{figure}
	\centering
	\includegraphics[width=0.9\linewidth]{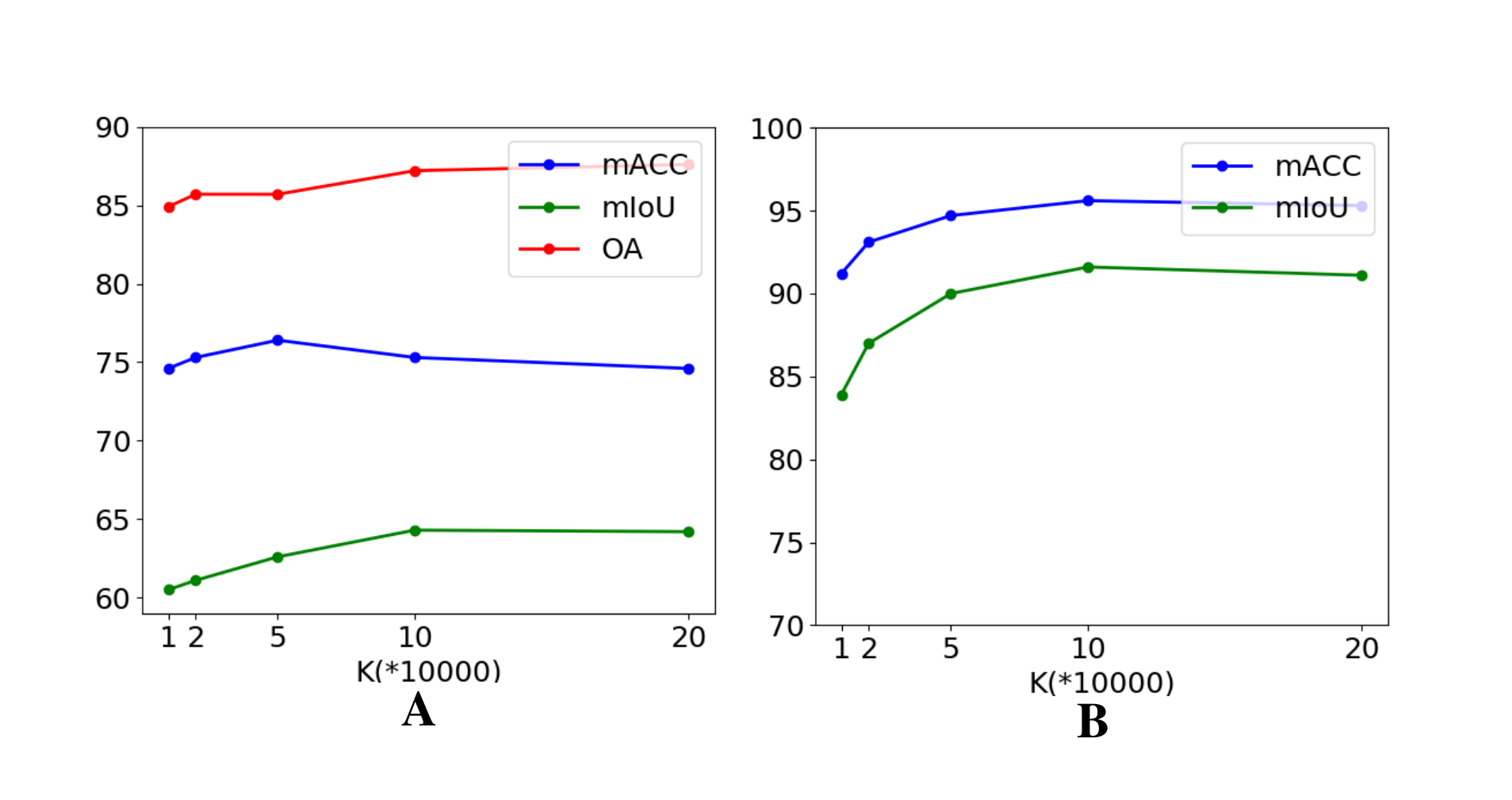}
	\caption{Performances (\%) under different numbers ($K$) of generated features. A for C3DS and B for Z3DS.}
	\label{fig4}
\end{figure}

We investigate the effect of the number ($K$) of generated features on the segmentation performance by conducting both C3DS and Z3DS experiments using RandLA-Net in the Area-$5$ validation setting on S3DIS. In C3DS, $K$ of each class is set to $\{10000, 20000, 50000, 100000, 200000\}$. In Z3DS, $K$ of each unseen class is set to $\{10000, 20000, 50000, 100000, 200000\}$, and the Z3DS task is performed under the 10/2 split. Figure~\ref{fig4}-A and Figure~\ref{fig4}-B show the results of C3DS ($mIoU$, $mACC$, and $OA$) and Z3DS ($mIoU$, $mACC$) respectively. As seen from Figure~\ref{fig4}, with the increase of the number of generated features, the segmentation performance increases accordingly at the beginning stage. This is reasonable because with more diverse point features, the trained point classifier can be more discriminative. When the number of generated features reaches a certain scale, the segmentation performance becomes relatively stable.

\subsubsection{Sensitiveness to Hyper-parameter($\gamma$)}
\begin{figure}
	\centering
	\includegraphics[width=0.9\linewidth]{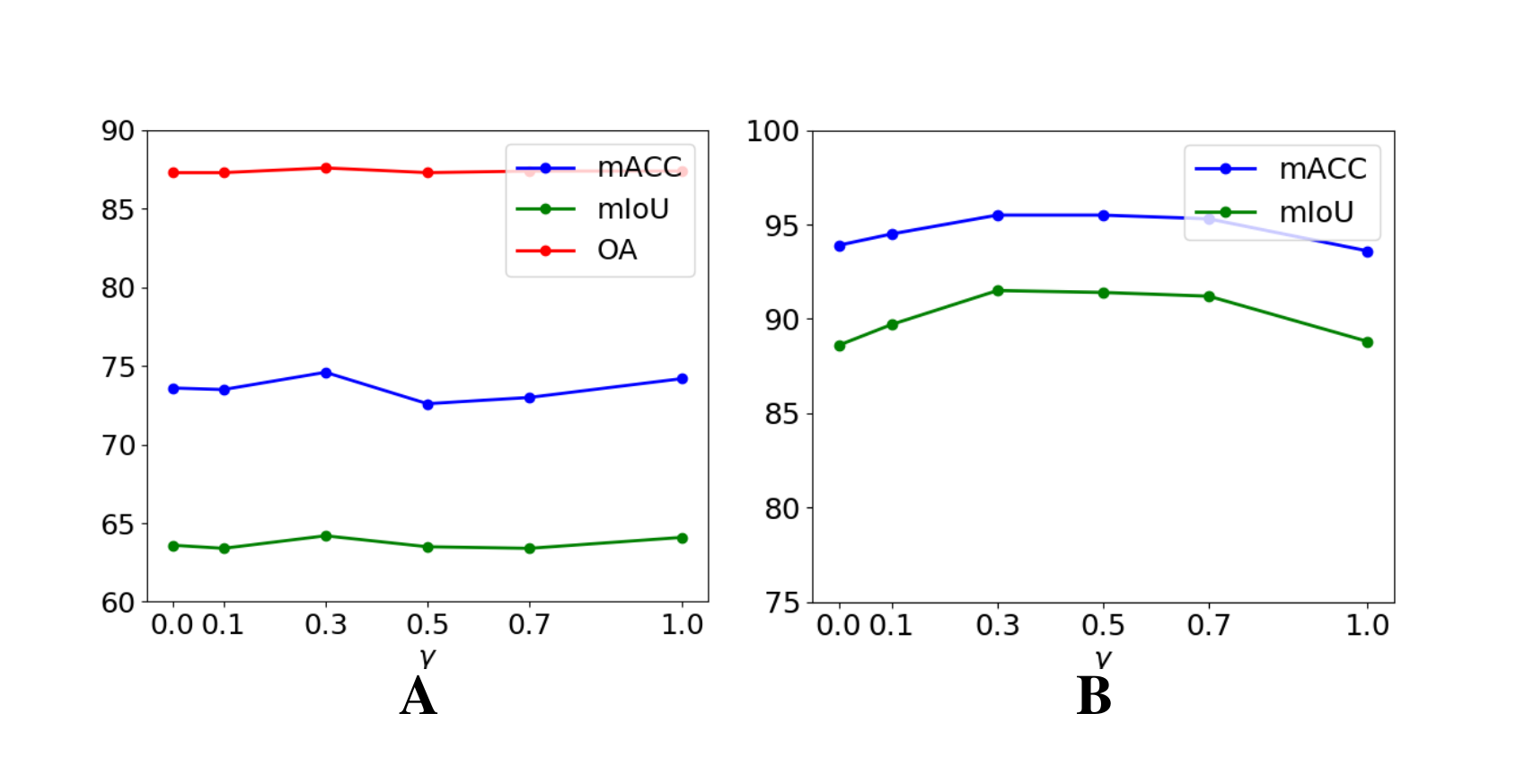}
	\caption{Performances (\%) under different ratio ($\gamma$) of the interpolated samples to the real samples. A for C3DS and B for Z3DS.}
	\label{fig5}
\end{figure}

Here we analyze the sensitiveness of the proposed framework to its key hyper-parameter, i.e. the ratio of the interpolated samples to the real samples ($\gamma$). We conduct both C3DS and Z3DS experiments using RandLA-Net as backbone network in the Area-$5$ validation setting on S3DIS, with $\gamma =\{0, 0.1, 0.3, 0.5, 0.7, 1.0\}$ respectively. The results of C3DS ($mIoU$, $mACC$, $OA$) and Z3DS ($mIoU$, $mACC$) are shown in Figure~\ref{fig5}-A and Figure~\ref{fig5}-B respectively. From Figure~\ref{fig5}, we can see that a moderate $\gamma$ can improve the performance, while a too large or too small $\gamma$ decreases the improvement. This is because a too small $\gamma$ can not take advantages of feature-geometry-based mixup training, while a too large $\gamma$ would make the interpolated samples overwhelming the real samples, hence obfuscating the decision boundary of classifier.

\subsubsection{Visualization}

\begin{figure}[t]
	\centering
	\includegraphics[width=0.95\linewidth]{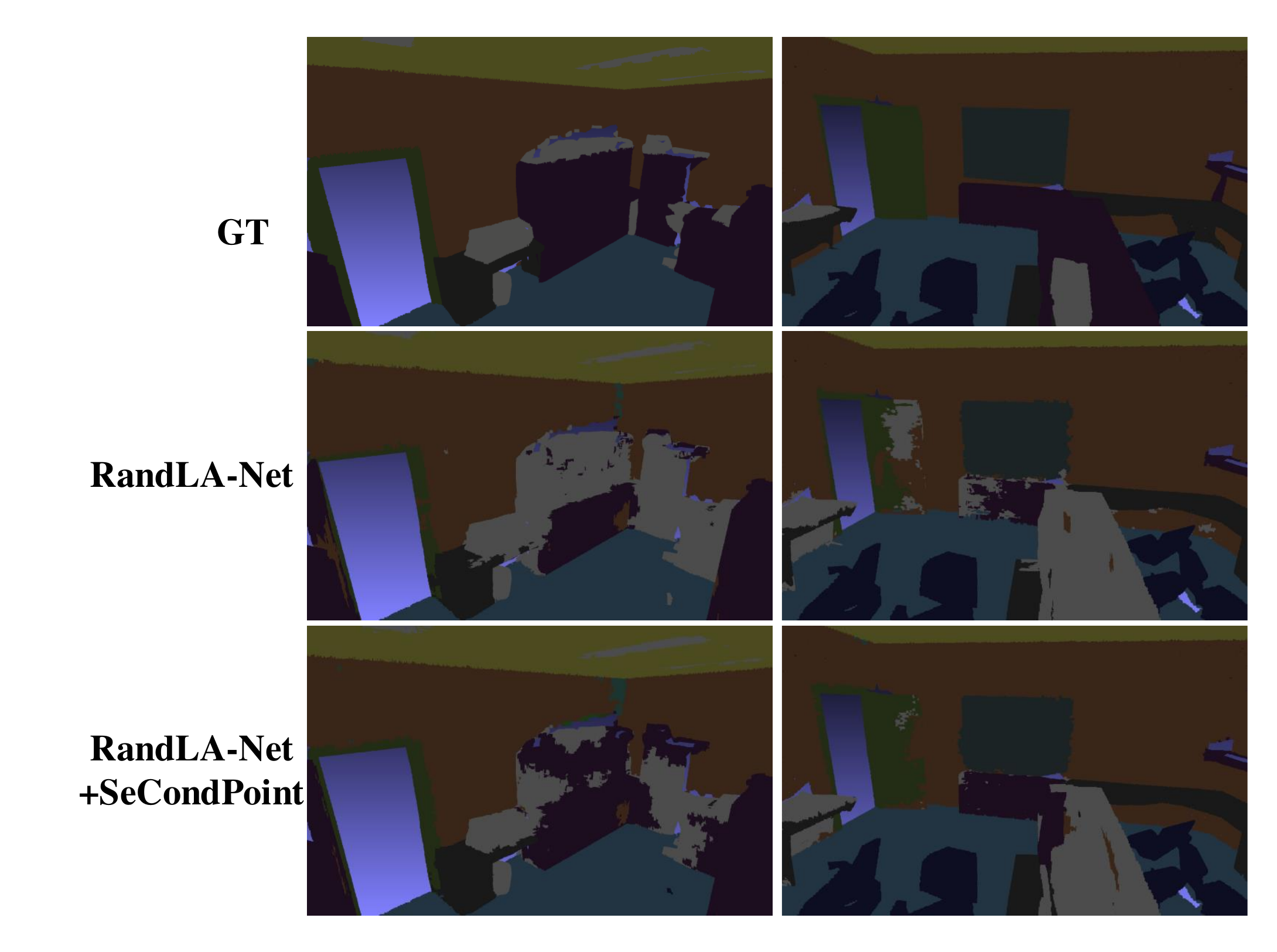}
	\caption{Visualization of the C3DS results of RandLA-Net and its enhanced version by the proposed SeCondPoint framework.}
	\label{fig6}
\end{figure}

\begin{figure}[t]
	\centering
	\includegraphics[width=0.95\linewidth]{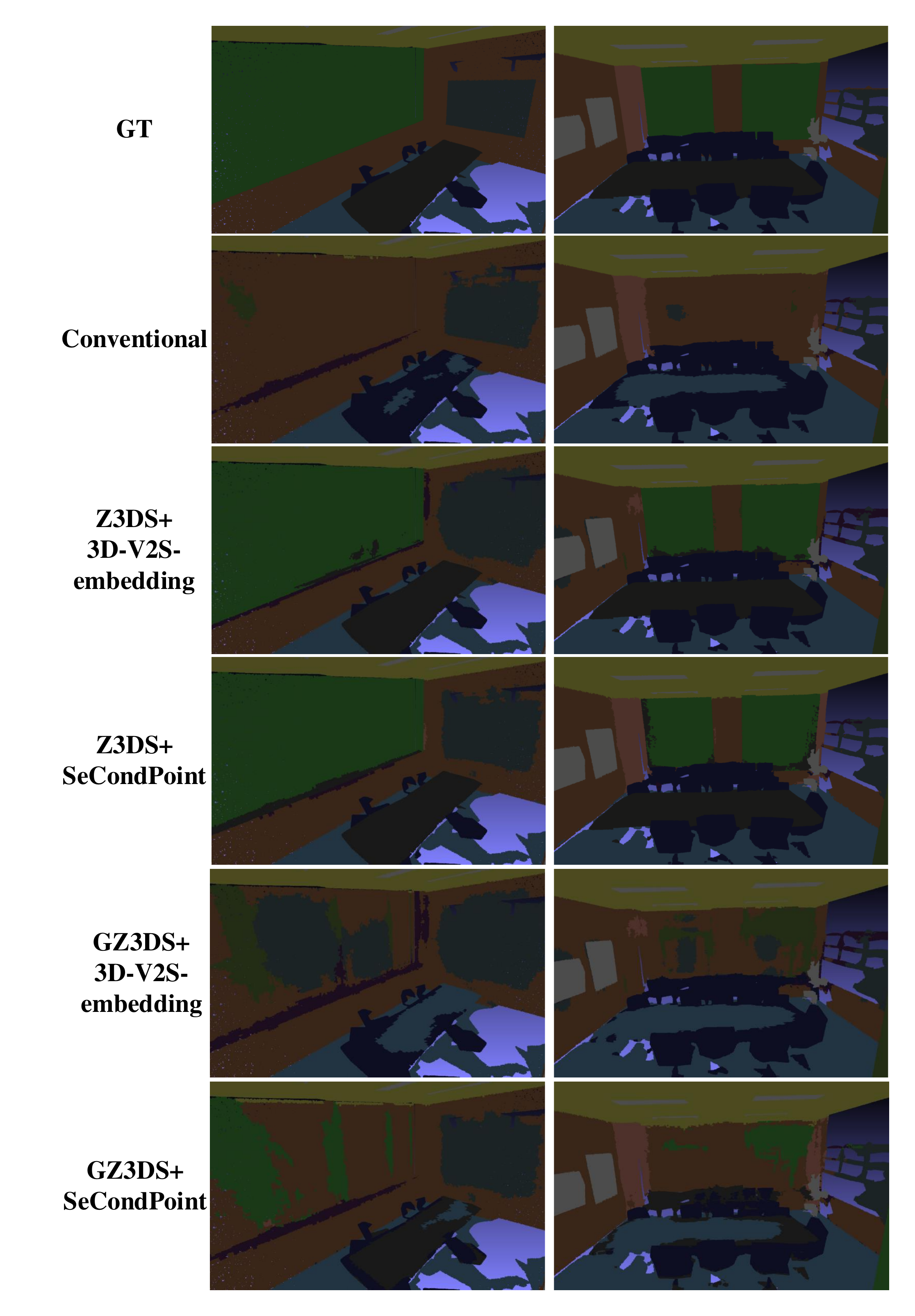}
	\caption{Visualization of the Z3DS/GZ3DS results of the conventional segmentation method, the baseline 3D-V2S-embedding, and the proposed SeCondPoint framework.}
	\label{fig7}
\end{figure}

To qualitatively demonstrate the effectiveness of the proposed framework, we visualize some segmentation results of the proposed framework and the comparative methods in the C3DS and the Z3DS/GZ3DS experiments in the Area-$5$ validation setting on S3DIS. In C3DS, the used comparative method includes the RandLA-Net backbone. In Z3DS, the used comparative methods include conventional segmentation network (i.e. RandLA-Net trained with $10$ seen classes), 3D-V2S-embedding. The 10/2 seen/unseen split is employed in Z3DS. The results of C3DS and Z3DS/GZ3DS are shown in Figure~\ref{fig6} and Figure~\ref{fig7} respectively. As shown in Figure~\ref{fig6}, in C3DS, the bookcase (purple) in the left view and the door (yellow green) in the right view are more accurately classified by RandLA-Net+SeCondPoint, demonstrating the proposed framework can improve the segmentation performance of the backbone network. From Figure~\ref{fig7}, we can see that 1) the conventional segmentation model always wrongly classifies the unseen-class objects into seen classes since it has no ability to recognize the unseen-class objects; 2) in Z3DS, the unseen-class objects (window (green) and table (dark gray)) are well recognized by both the baseline 3D-V2S-embedding and the proposed SeCondPoint, demonstrating that the introduced language-level semantics are effective at novel-class inference; and 3) in GZ3DS, the segmentation of unseen classes is significantly worse than that in Z3DS for both 3D-V2S-embedding and our proposed method, showing the severe bias problem. However, compared with 3D-V2S-embedding, the proposed framework can significantly alleviate the bias problem, successfully inferring a large number of novel-class (window (green) and table (dark gray)) points.

\section{Conclusion and Future Work}
\label{conclusion}
In this paper, we propose a unified SeCondPoint framework for both conventional 3D scene semantic segmentation (C3DS) and zero-shot 3D scene semantic segmentation (Z3DS), where language-level semantics are introduced to condition the modeling of point feature distribution and a large number of point features from both seen and unseen classes can be generated from the learned distribution. The proposed framework has the flexibility to not only enhance the performances of existing segmentation networks by augmenting their feature diversity, but also endow the existing networks with the ability to handle unseen-class objects. The C3DS experimental results on two public datasets demonstrate that the proposed framework is able to significantly boost the performances of existing segmentation networks. Two benchmarks are also introduced for the research of the newly introduced Z3DS task, and the results on them demonstrate the effectiveness of the proposed framework. 

In the future, a number of directions are worthy of further exploring. At first, our newly introduced Z3DS task could be considered as a practical setting for 3D scene segmentation since novel classes are often encountered in realistic scenarios, then inspirations could be taken from recent advanced zero-shot learning methods and the characteristics of 3D scene data (such as geometrical knowledge), for example, the bias problem in GZ3DS is expected to be alleviated by resorting to some novel-class detectors. In addition, in the current semantics conditioned feature modeling framework, simple word embeddings are used as semantic features and GAN is employed to model feature distributions. In fact, under the proposed framework, semantics are not constrained to word embeddings. Higher-quality semantics, such as sentence embeddings which represent the descriptions of 3D objects and their spatial relationships, and semantic attributes which have explicit semantic meanings pre-defined by experts, could also be used to achieve better performances. Besides, the way to use language-level semantics is not necessarily constrained to modeling feature distribution with GAN. Other generative models like VAEs and NFs could also be used to model feature distributions, and other advanced embedding networks could be used to jointly learn point representations and semantic representations. In a word, the present work introduced a promising semantics conditioned framework, and its key ingredients are largely subject to much further improvements.


\section*{Acknowledgment}
This work was supported by the National Natural Science Foundation of China (NSFC) under Grants (61991423, U1805264) and the Strategic Priority Research Program of the Chinese Academy of Sciences (XDB32050100).

\ifCLASSOPTIONcaptionsoff
  \newpage
\fi

\bibliographystyle{IEEEtran}
\bibliography{ref}

\end{document}